\documentclass[sigconf]{acmart}
\def\BibTeX{{\rm B\kern-.05em{\sc i\kern-.025em b}\kern-.08emT\kern-.1667em\lower.7ex\hbox{E}\kern-.125emX}}
    
\copyrightyear{2020}
\acmYear{2020}
\setcopyright{acmcopyright}
\acmConference[ACMSE 2020]{2020 ACM Southeast Conference}{April 2--4, 2020}{Tampa, FL, USA}
\acmBooktitle{2020 ACM Southeast Conference (ACMSE 2020), April 2--4, 2020, Tampa, FL, USA}
\acmPrice{15.00}
\acmDOI{10.1145/3374135.3385287}
\acmISBN{978-1-4503-7105-6/20/03}

\def\BibTeX{{\rm B\kern-.05em{\sc i\kern-.025em b}\kern-.08emT\kern-.1667em\lower.7ex\hbox{E}\kern-.125emX}}
    
\usepackage{flushend} 
\usepackage{geometry}
\hypersetup{draft}

\begin{document}

%
\title{Risk Prediction of Peer-to-Peer Lending Market by a LSTM Model with Macroeconomic Factor}

%
\author{Yan Wang}

\orcid{1234-5678-9012}
\affiliation{%
  \institution{Kennesaw State University}
  \city{Kennesaw, GA}
  \state{USA}
}
\email{ywang63@students.kennesaw.edu}

\author{Xuelei Sherry Ni}
\affiliation{%
  \institution{Kennesaw State University}
  \city{Kennesaw, GA}
  \state{USA}
}
\email{sni@kennesaw.edu}


 




%
\renewcommand{\shortauthors}{Wang and Ni.}

%
\begin{abstract}
In the peer to peer (P2P) lending platform, investors hope to maximize their return while minimizing the risk through a comprehensive understanding of the P2P market. 
A low and stable average default rate across all the borrowers denotes a healthy P2P market and provides investors more confidence in a promising investment.
Therefore, having a powerful model to describe the trend of the default rate in the P2P market is crucial. 
Different from previous studies that focus on modeling the default rate at the individual level, in this paper, we are the first to comprehensively explore the monthly trend of the default rate at the aggregative level for the P2P data from October 2007 to January 2016 in the US.
We use the long short term memory (LSTM) approach to sequentially predict the default risk of the borrowers in Lending Club, which is the largest P2P lending platform in the US. 
Although being first applied in modeling the P2P sequential data, the LSTM approach shows its great potential by outperforming traditionally utilized time series models in our experiments.
Furthermore, incorporating the macroeconomic feature \textit{unemp\_rate} (i.e., unemployment rate) can improve the LSTM performance by decreasing RMSE on both the training and the testing datasets. 
Our study can broaden the applications of the LSTM algorithm by using it on the sequential P2P data and guide the investors in making investment strategies.
\end{abstract}

%

%

%
\maketitle
\section{Introduction}
Peer to peer (P2P) lending, which means lending money from investors directly to borrowers through a virtual platform, is one of the fastest-growing segments in the financial lending market.
Through the P2P lending platform, approved borrowers could take control of their finance while investors benefit via earning potentially competitive returns \cite{bachmann2011online}. 
To help the investors make the investment decision, 
the lending institutions continuously focus on exploring methods to understand the behavior of loan applicants during the economic cycles.
They attempt to model the default risk of the borrowers (i.e., repayment of the loans) and then provide credit assessment to the lenders \cite{kim2019predicting}.
Meanwhile, it is equally important for the investors to have a whole understanding of the entire P2P market by evaluating the borrowers' risk at the aggregate level as time going on.
Lending platform with a continuously low and stable default risk may denote a healthy P2P lending environment, thus could provide more confidence to the investors to have a successful investment \cite{emekter2015evaluating}\cite{mariotto2016competition}\cite{chen1970comparative}.
Therefore, how to model the trend of the default risk at the aggregative level becomes a critical question that needs to be addressed. 

The long short term memory (LSTM) model, which is one of the state-of-the-art methods to model the sequential data (i.e., the order of the data matters), has been widely used in language modeling, disease forecasting, and speech recognition \cite{sundermeyer2012lstm}\cite{liu2018lstm}\cite{jagannatha2016bidirectional}\cite{graves2013hybrid}.
In the financial domain, LSTM has shown its superiority over traditional time series models in individual credit risk classification, overdue of bank loan prediction, and credit card fraud detection \cite{jurgovsky2018sequence}\cite{li2018overdue}\cite{wang2018deep}. 
Although the LSTM model has been applied to the above-mentioned fields, no research has been found to analyze the time series data in the aggregative level generated in the P2P lending market. 
It is worth noting that different from modeling on the individual repayment that focuses on individual characteristics, modeling on the aggregative data will also need to consider the macroeconomic factors that are relevant to the P2P market.
For example, one macroeconomic factor, the unemployment rate, is shown to be closely correlated to the interest rate in the P2P lending market \cite{foo2017macroeconomics}.
Moreover, the unemployment rate is empirically correlated with gross domestic product (GDP) \cite{sogner2001okun}. 
All these findings show strong evidence that we need to incorporate the macroeconomic factors when modeling the default rate in the P2P market. 

Motivated by the aforementioned research, in this paper, we demonstrate a comprehensive case study with the goal to model the trend of the default rate of the P2P market at the aggregative level in the US.
In our empirical study, we use the Lending Club data to test the robustness of LSTM.
We first combine the P2P data from the individual level to the aggregative level. 
Next, we incorporate the employment rate (i.e., $unemp\_rate$) across different time points into the aggregated data by matching the date. 
Then, the LSTM model is employed to fit the aggregated default rate.
The superiority of the LSTM method is confirmed by comparing its performance to traditional time series models. 
Furthermore, the importance of the macroeconomic factor is proved by comparing the performance of the LSTM models with or without $unemp\_rate$.
The authors believe that our findings could provide a reference from the aggregative level for the investors for making their decisions. 
In summary, our study makes contributions from the following aspects:
\begin{itemize}
\item It is the first attempt that utilizes LSTM on the aggregated sequence data in P2P lending. The LSTM model is shown to be superior than the traditionally used time series models; 

\item It is the first attempt that incorporates the macroeconomic factor named unemployment rate into LSTM modeling for the repayment prediction at the aggregative level. 
We found that adding the macroeconomic factor is beneficial to the model performance.
\end{itemize}

The rest of the paper is structured as follows. 
Section \ref{relatedwork} summarizes the related work in the P2P lending market. 
Section \ref{algorithms} provides a description of the LSTM algorithm along with its origin algorithm -- recurrent neural network (RNN). Section \ref{methodology} introduces the details of our analysis and Section \ref{result} presents the results.
Section \ref{conclusion} is the conclusion of our study.

\section{Related Work} \label{relatedwork}
Many previous studies focus on exploring different machine learning algorithms to model the repayment of individual borrowers in P2P lending.
They employ different models such as random forest, decision tree, and neural network, to improve the classification accuracy for loan status or to extract efficient features that are predictive of default \cite{malekipirbazari2015risk}\cite{namvar2018credit}\cite{ye2018loan}. 
These studies could guide the investment strategies for the investors by providing evaluations of individual borrowers.
However, there is seldom research that describes the sequential development of the default risk in the P2P market as time going on.
In other words, there is no research that could provide a reference to the investors on the overall evaluation of the default risk at the aggregative level in the P2P market.
A deep learning approach has been explored in many other areas for modeling the sequential data.
For example, RNN has been introduced into the Internet recommendation system for building a recommendation system in \cite{hidasi2015session}.
LSTM has shown to be effective in the prediction of the future behavior of customers in the e-commerce based data in \cite{molina2018understanding}. 
In addition, LSTM has shown its superior over the traditionally utilized time series models when being applied to model the transaction fraud and credit scoring in \cite{wiese2009credit}\cite{wang2018deep}.
Although not been applied in modeling the sequential P2P data at the aggregative level, LSTM is expected by us to have its potential.
Thus, we did an empirical study to confirm our conjecture and details of our study will be discussed in Section \ref{methodology}.

\section{Algorithms} \label{algorithms}
Since the LSTM model is used in this study, in this section, we will first briefly discuss RNN, which is the origin of LSTM, and then illustrate the principle of LSTM.

\subsection{Recurrent Neural Network}
\begin{figure}[htbp]
	\centering
	\includegraphics[width=3.3in]{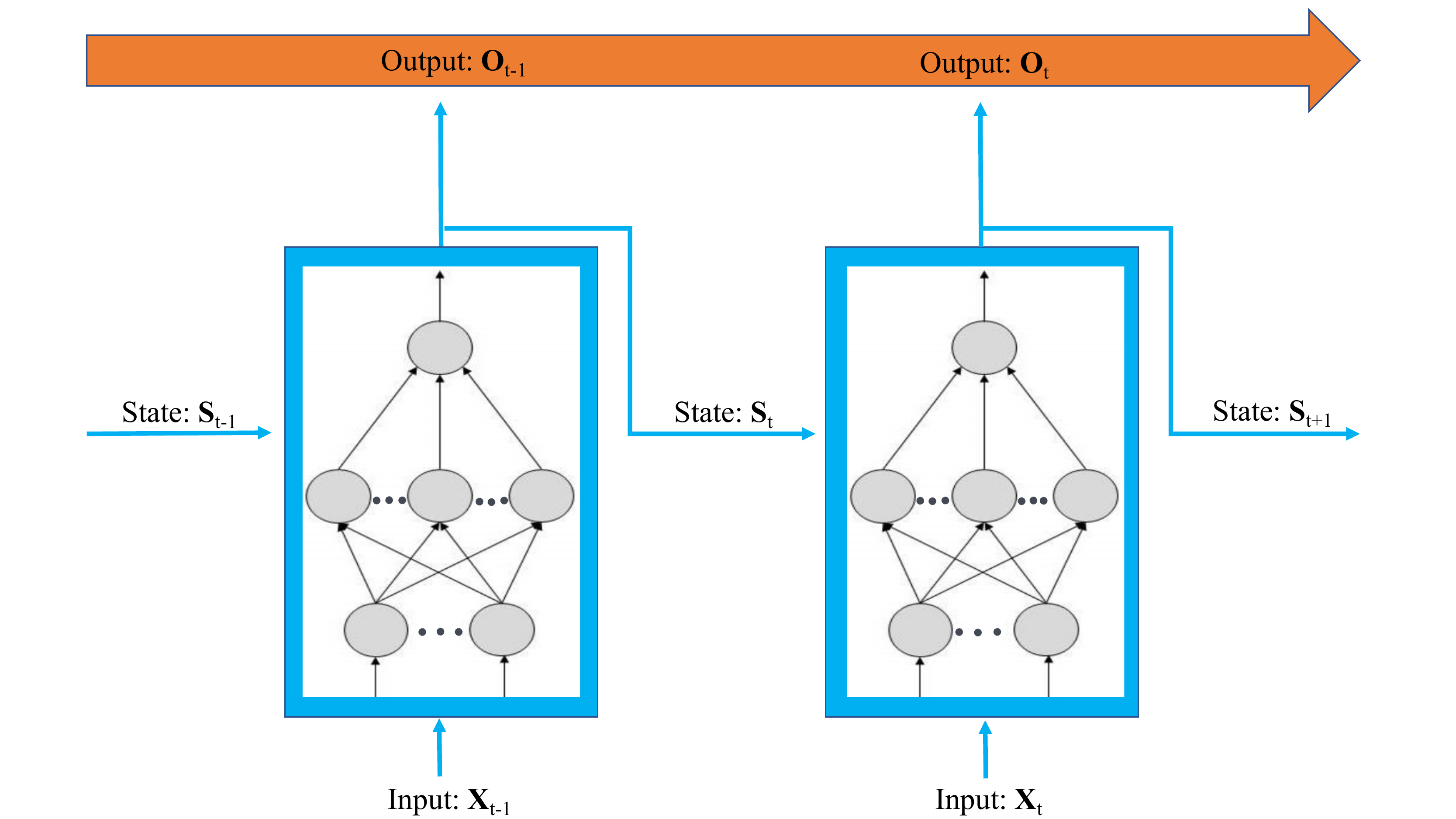}
	\caption{Illustrative Figure for an Example of RNN Structure}
	\label{RNN_final}
\end{figure}

In traditional feed-forward neural networks (NNs), the information of the data moves towards one direction: from the input layer, through the hidden layer(s), and finally reaches the output layer. 
Thus, NNs only store the current information they received and have no memory of the past. 
As a result, they have limited power when being used on sequential data such as transaction data or speech data \cite{jain1996artificial}. 
On the other hand, RNN, a special class of NNs, has shown its potential in modeling data with temporal dynamic behavior by many studies \cite{graves2013speech} \cite{ho2002comparative}.
Different from NNs, data information cycles in RNN and the current information along with the previous step information can both be stored.
In other words, RNN has the internal while short-term memory of the information that NNs do not \cite{le2015tutorial}. 
Figure \ref{RNN_final} displays an illustrative example of an RNN structure.
Each rectangle denotes a fully-connected NN structure (note: the structure in Figure \ref{RNN_final} is shown as an illustrative example and the exact NN structure needs to be self-defined in different studies) and the RNN is composed of a chain of repeating the same NN structure.
At each timestamp t, besides using the values of the independent variables at time t (i.e., $\textbf{X}_t$) as the input, RNN also uses the output from the previous timestamp (i.e., $\textbf{S}_t$) as the input. 
The output at time t of RNN (i.e., $\textbf{O}_{t}$) can be calculated using Equation \ref{RNN_equation}, where ``$\cdot$" denotes the Hadamard product (i.e., pointwise multiplication), $activation$ denotes a certain activation function (such as sigmoid function), $W$ and $U$ denote the weight matrix for $\textbf{X}_t$ and $\textbf{S}_t$, and $\textbf{b}$ denotes the bias. 
By doing this, `memory' could be added on RNN and the sequential information of the data is stored as time goes on.
It is worth noting that in RNN, values of $\textbf{O}_{t}$ and $\textbf{S}_{t+1}$ are the same for each time point $t$, with the former denotes the current output and the latter represents the information passing to the next time point $t+1$.

\begin{equation} \label{RNN_equation}
\textbf{O}_{t} = activation(\textbf{W} \cdot \textbf{X}_t + \textbf{U} \cdot \textbf{S}_t + \textbf{b})
\end{equation}

\subsection{Long Short Term Memory}
\begin{figure}[h]
	\centering
	\includegraphics[width=3.3in]{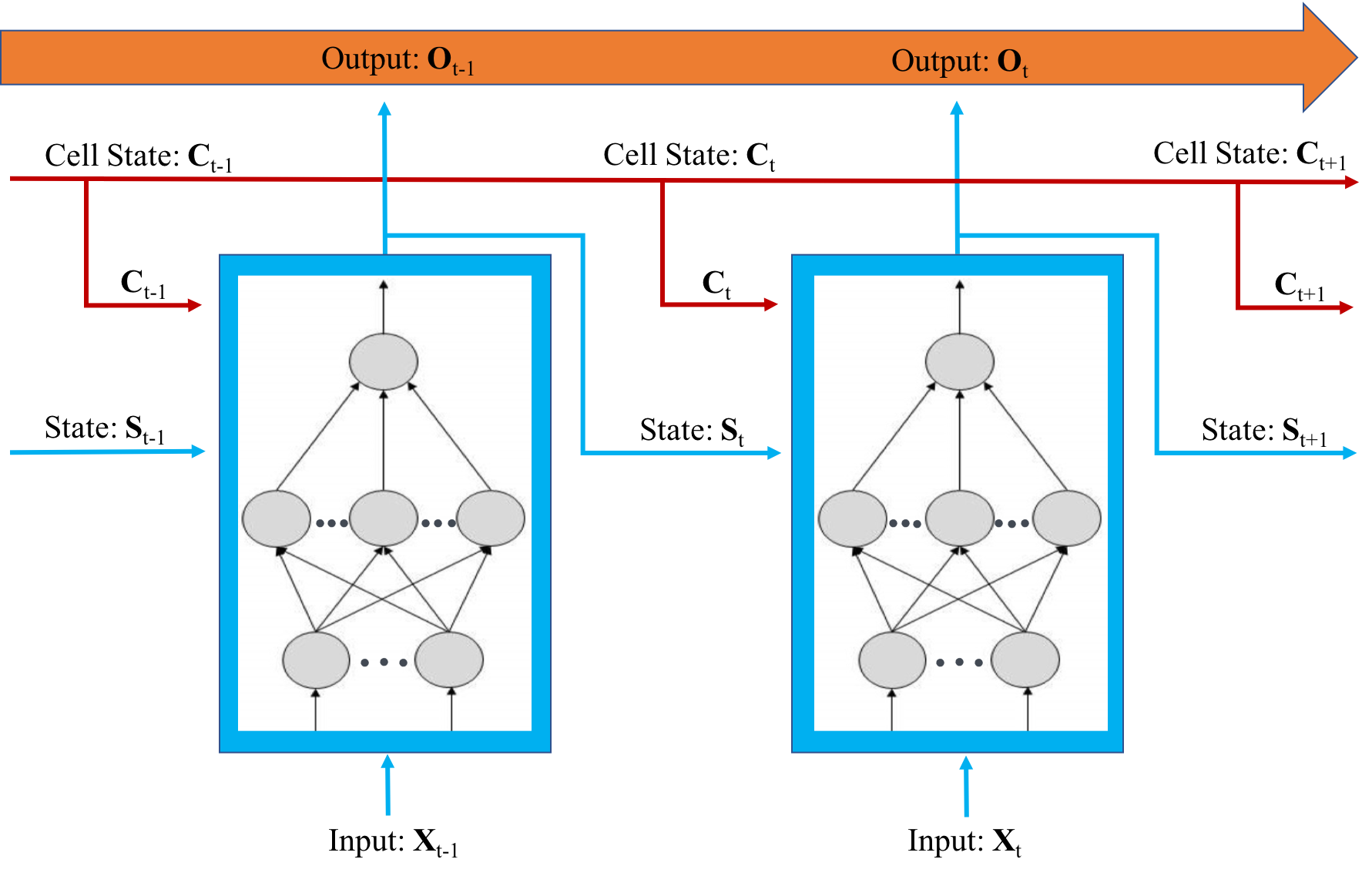}
	\caption{Illustrative Chart for an Example of LSTM Structure}
	\label{LSTM_final}
\end{figure}

LSTM is a variant of RNN but it is capable of remembering the information over a long period of time and learning long-term dependencies of the information.
In other words, it extends the `memory' and could learn from inputs that have a very long time lags in between. 
Figure \ref{LSTM_final} displays an illustrative example of a LSTM structure. Comparing with the RNN structure in Figure \ref{RNN_final}, it is found that in Figure \ref{LSTM_final}, LSTM contains an additional sequence of cell states {$\textbf{C}_{t}$},
which not only stores the previous information, but also the information obtained many steps ahead.
Similarly, the output of LSTM at time t (i.e., $\textbf{O}_{t}$ in Figure \ref{LSTM_final}) can be calculated using Equation \ref{LSTM_equation}, where $\textbf{W}_{o}$, $\textbf{U}_{o}$, and $\textbf{V}_{o}$ denote the corresponding weight matrix (for $\textbf{X}_t$, $\textbf{S}_t$, and $\textbf{C}_t$ respectively), and $b_{o}$ denotes the bias \cite{hammer2002tutorial}.
Similar as those in RNN of Figure \ref{RNN_final}, values of $\textbf{O}_{t}$ and $\textbf{S}_{t+1}$ are the same for each time point $t$, with the former denotes the current output and the latter represents the information passing to the next time point $t+1$.

\begin{equation} \label{LSTM_equation}
\textbf{O}_{t} = activation(\textbf{W}_{o} \cdot \textbf{X}_t + \textbf{U}_{o} \cdot \textbf{S}_t + \textbf{V}_{o} \cdot \textbf{C}_{t} + \textbf{b}_{o})
\end{equation}

The critically innovative structure of LSTM is the cell state $\textbf{C}_{t}$. 
Its detailed structure is summarized based on the illustrations from previous studies \cite{hochreiter1997lstm}\cite{ren2016look}. 
As shown in Figure \ref{LSTM_final}, the sequence of cell states is similar to a conveyor belt or a carry track that controls whether to input, store, or delete information. 
For each cell state, it contains different gates that could optionally delete, store, or output information: the forget gate $\textbf{f}_t$ (particially) deletes the information from previous state if it is not important, the input gate $\textbf{i}_t$ determines the percentage of new input, and the output gate $\textbf{O}_t$ denotes the output at the current time step $t$.
$\textbf{f}_t$ can be obtained by using Equation \ref{forget}, where $\textbf{W}_f$ and $\textbf{U}_f$ denotes the weight matrix for $\textbf{X}_t$ and $\textbf{S}_t$ of the forget gate, and $\textbf{b}_f$ denotes the bias.
Similarly, $i_t$ can be obtained by using Equation \ref{input1}, where $\textbf{W}_i$ and $\textbf{U}_i$ denotes the weight matrix for $\textbf{X}_t$ and $\textbf{S}_t$ of the input gate, and $\textbf{b}_i$ denotes the bias.
After obtaining the information that passing through the input gate (i.e., $\textbf{i}_t$), LSTM uses another layer to generate a new candidate value $\tilde C$, which denotes the information that could be added to the current state $\textbf{C}_t$. 
The candidate value $\tilde C$ can be obtained by using Equation \ref{input2}, where $\textbf{W}_k$ and $\textbf{U}_k$ denotes the weight matrix for $\textbf{X}_t$ and $\textbf{S}_t$, and $\textbf{b}_k$ denotes the bias.
Finally, the current cell state $\textbf{C}_t$ can be updated into the new cell state $\textbf{C}_{t+1}$ by using Equation \ref{ct}, where * denotes the matrix multiplication, $\textbf{f}_t$*$\textbf{C}_t$ denotes the information LSTM wants to delete at time $t$, and $\textbf{i}_t$*$\tilde C$ denotes the information LSTM wants to remain. 
Then, $C_{t+1}$ would be used to calculate the output in the next time step $t+1$ (i.e., $\textbf{O}_{t+1}$) \cite{lipton2015critical}.

\begin{equation} \label{forget}
\textbf{f}_t = activation(\textbf{W}_f \cdot \textbf{X}_t + \textbf{U}_f \cdot \textbf{S}_t + \textbf{b}_f)
\end{equation}

\begin{equation} \label{input1}
\textbf{i}_t = activation(\textbf{W}_i \cdot \textbf{X}_t + \textbf{U}_i \cdot \textbf{S}_t + \textbf{b}_i)
\end{equation}

\begin{equation} \label{input2}
\tilde C = activation(\textbf{W}_k \cdot \textbf{X}_t + \textbf{U}_k \cdot \textbf{S}_t + \textbf{b}_k)
\end{equation}

\begin{equation} \label{ct}
\textbf{C}_{t+1}= \textbf{f}_t * \textbf{C}_t + \textbf{i}_t * \tilde C
\end{equation}

During the training of LSTM, the sequential cell states (examples including $\textbf{C}_{t-1}$, $\textbf{C}_{t}$, and $\textbf{C}_{t+1}$) are trained at a series of time points (including $t-1$, $t$, and $t+1$) by identifying the optimal weights and bias with the goal of minimizing the pre-defined loss function.  
As mentioned above, since each square in Figure \ref{RNN_final} and \ref{LSTM_final} denotes a fully-connected NN structure, LSTM contains similar hyper-parameters as traditional NNs, such as `number of nodes', `batch\_size' (i.e., number of samples used for propagation in each iteration), and `number of epochs' (i.e., number of times that the learning algorithm sees the entire dataset). 
The optimal values of the hyper-parameters need to be identified for different datasets before starting the model training.

\section{Empirical Study} \label{methodology}
\subsection{Dataset} \label{data}
The empirical study in this paper uses the Lending Club data downloaded via the website
\footnote{\textit{https://www.lendingclub.com/info/download-data.action}}.
The dataset records the P2P lending transactions from Lending Club (which is the largest US P2P lending platform) ranging from 2007 to 2017.
Since Lending Club is the largest lending platform in the US, the data is a good representative of the entire P2P market in the US. 
There are millions of loan transactions and each transaction is identified by the unique ID.
For each transaction, there are over thirty features that describe the financial information of the money borrowers as well as the information related to the loan such as the starting date, the amount of the loan, and the term of the loan. 
The variable \textit{loan\_status} describes the different status of the loan transactions: ongoing, fully paid off, or default.
In our study, we remove the loan cases that are still ongoing.
As a result, the target variable \textit{loan\_status} retains two categories: fully paid off or default.

The features in the dataset mainly fall into three categories: personal property (PP), credit information (CI), or loan information (LI). 
Table \ref{featuredc} provides the descriptions, types, as well as the categories of the retained variables after we removing those with ambiguous meanings. 
Except for the target variable \textit{loan\_status}, most features are numerical and there are only three categorical features. 
It is worth noting that since our analysis will be based on the aggregative level, it is critical to explore some macroeconomic factors in addition to the individual factors. 
This concern has been proved by many previous research, which showed the potential effect of the macroeconomic behavior on \textit{loan\_status} such as unemployment rate and S\&P500 index \cite{jakubik2007macroeconomic}. 
In our analysis, we collect one macroeconomic feature using the website
\footnote{\textit{https://datahub.io/core/employment-us\#data.}}.
The feature is named as \textit{unemp\_rate} and it is recorded monthly. 
It will be served as an additional numerical feature in the following analysis. 

\begin{table*}[htbp]
\renewcommand{\arraystretch}{1.3}
\caption{Variables Kept in the P2P Lending Transaction Dataset}
\label{featuredc}
\centering
\begin{tabular}{|p{35mm}|p{95mm}|p{15mm}|p{15mm}|}

			\hline
			Feature name & Description & Category & Type \\
			\hline
			$application\_type$  & Indicates whether the loan is an individual application or a joint application with two co-borrowers & LI & Categorical \\
			$home\_ownership$  & Home ownership status of the borrowers & PP & Categorical \\
			$varification\_status$  & Indicates if income was verified by LC, not verified, or if the income source was verified & PP & Categorical \\
			$loan\_status$ (target) & The loan is fully paid off or default & LI & Binary \\
			$annual\_inc$ & Annual income reported by the borrowers & PP & Numerical \\
			$collection\_recovery\_fee$ & Post charge off collection fee & LI & Numerical \\
			$delinq\_amnt$ & The past-due amount owed for the accounts on which the borrower is now delinquent & CI & Numerical \\
			$delinq\_2yr$  & Number of over 30 days past-due incidences of delinquency in the borrowers' credit files for the past 2 years & CI &Numerical \\
			$int\_rate$  & Interest rate on the loan & LI & Numerical \\
			$installment$  & The monthly payment owed by the borrower if the loan originates & LI & Numerical \\
			$last\_pymnt\_amnt$  & Last total payment amount received & LI & Numerical \\
			$loan\_amnt$  & The amount of the loan & LI & Numerical \\
			$open\_acc$  & Number of accounts opened in past 24 months & CI & Numerical \\
			$pub\_rec$  & Number of derogatory public records & CI & Numerical \\
			$recoveries$  & Post charge off gross recovery & LI & Numerical \\
			$revol\_bal$  & Total credit revolving balanced & CI & Numerical \\
			$total\_acc$  & The total number of credit lines in the borrower's credit file & CI & Numerical \\
			$total\_pymnt$  & Payments received to date for total amount funded & LI & Numerical \\
			$total\_rec\_late\_fee$ & Late fees received to date & LI & Numerical \\
				\hline
		\end{tabular}
\end{table*}

\subsection{Data Pre-processing} \label{processing}
Before applying the LSTM algorithm, several data pre-processing procedures are performed as follows:

(a) Remove redundant information:
With respect to the target variable \textit{loan\_status}, as mentioned in Section \ref{data}, observations with \textit{loan\_status} valued `ongoing' are removed.
With respect to the features (both numerical and categorical), those having missing/invalid percentage larger than $80\%$ were removed.
We then transform the target variable \textit{loan\_status} to numerical by giving the observations with \textit{loan\_status} valued `fully paid off' a value `0' while those valued `default' a value `1'.
As a result, there remain around one million observations and the transaction time ranges from October 2007 until January 2016. 

\begin{figure}[htbp]
	\centering
	\includegraphics[width=3.3in]{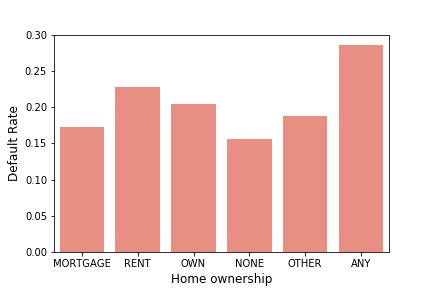}
	\caption{Default Rate in the Borrowers with Different Categories of Home Ownership}
	\label{categoryEDA}
\end{figure}

(b) EDA on categorical features: Exploratory data analysis (EDA) is implemented with the goal to first understand the distribution of each categorical feature and then to determine whether we should pool different categories of a variable together. 
As described in Table \ref{featuredc}, except for the target variable, there are only three categorical features in the dataset:
\textit{home\_ownership}, \textit{verification\_status}, and \textit{application\_type}. 
Take \textit{home\_ownership} as an example to show our data pre-processing step on the categorical features. 
Figure \ref{categoryEDA} displays the percentage of delinquency (i.e., the percentage of \textit{loan\_status} = 1) in each level of \textit{home\_ownership}.
The Wilcoxon rank-sum test shows that at a significant level of 0.05, there is a statistically significant difference in the percentage of default among the six different levels of \textit{home\_ownership}. 
Therefore, we keep all these six levels and use the one-hot-encoding method to convert each category into numerical values \cite{potdar2017comparative}.
Similar strategies are applied to \textit{verification\_status} (including three levels: `not verified', `verified', and `source verified') and \textit{application\_type} (including two levels: `individual' and `joint application').

(c) Missing value imputation: 
For the three categorical features shown in Table \ref{featuredc}, they have all been coded into numerical values after the one-hot-encoding transformation. Their missing values are imputed using the mode values. 
For the numerical features shown in Table \ref{featuredc}, median-based imputation is applied.

(d) Transfer data into the aggregative level:
We aggregate the data by month to get the sequentially monthly information of the P2P market.
The details of our aggregation are described as follows. 
(1): For the target variable \textit{loan\_status}, we calculate the percentage of \textit{loan\_status} = 1 within each month and use this as the aggregated value.
As a result, we obtain the monthly default rate of the P2P lending market and we name it as \textit{default\_rate} in the further analysis;
(2): For the independent variables, they all have been transformed into numerical values as mentioned in steps (b) and (c). 
Therefore, the monthly aggregated values are obtained by taking the monthly average for each feature.

(d) Append the macroeconomic factor: The monthly values of \textit{unemp\_rate} is finally merged with the aggregated Lending Club data by matching the date.

\begin{figure}[htbp]
	\centering
	\includegraphics[width=3.3in]{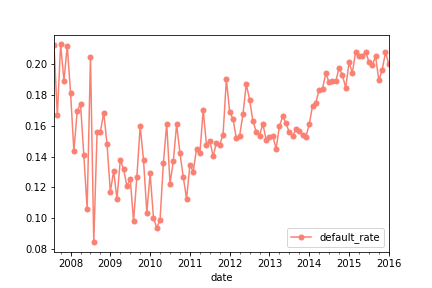}
	\caption{Monthly Change of the Default Risk at the Aggregative Level}
	\label{numvar}
\end{figure}

\subsection{Prediction of Default Risk} \label{RNN}
\subsubsection{LSTM}
After the aforementioned data pre-processing, we obtain 102 observations on the aggregative level along with 20 variables (18 independent variables from the original data, 1 macroeconomic factor, and 1 dependent variable). 
We plot the trend of the \textit{default\_rate} at the aggregative level using the line plot and the result is displayed in Figure \ref{numvar}.
It is observed that \textit{default\_rate} gradually decreases from October 2007 to early 2010 with big variations but it begins to increase afterward and the variation becomes smaller. 
The LSTM approach is applied to model the aggregated sequential default rate. 
The dataset obtained from Section \ref{processing} was split into a $80\%$ training and $20\%$ testing. 
To be specific, we use the data from October 2007 to May 2014 as the training set while using that from June 2014 to January 2016 as the testing set. 
The implementation of the LSTM model is based on the Keras library in Python 3 on a personal laptop with a 3.3 GHz Intel Core i7 processor, 16GB RAM, and Mac OS system. 
The loss function used in LSTM is the square root of mean squared error (RMSE) between the predicted and the true default rate \cite{marino2016building}. 
During the training process, we tuned several hyper-parameters of LSTM including `number of nodes', `batch\_size', and `number of epochs' via a trial and error approach with the goal of minimizing the cross-validated RMSE.
We keep the default settings in the Keras library for the rest of the hyper-parameters in LSTM.

To identify whether the incorporated macroeconomic feature, \textit{unemp\_rate}, is beneficial to the mode performance, two LSTM models are implemented as follows: (I) the LSTM model without using \textit{unemp\_rate}, denoted as LSTM(1); (II) the LSTM model by using \textit{unemp\_rate} as an additional feature, denoted as LSTM(2). 
It is worth noting that LSTM is relatively robustness to the multicollinearity problem \cite{zohrevand2017deep}, making us confident to use all the features simultaneously in the modeling stage.

\subsubsection{Further Comparison} \label{TTS}
To further explore the robustness and superiority of the LSTM technique in modeling the sequential default rate of the P2P lending data, traditional time series analysis is applied and evaluated on the same dataset described in Section \ref{RNN}. 
In our initial analysis, we considered both the univariate time series model (UTS, i.e., \textit{default\_rate} depends only on time) and multivariate time series model (MTS, i.e., \textit{default\_rate} depends on several time-dependent variables) \cite{zhang2017time} \cite{xiong2017detrended}. 
In UTS, the plots of the auto-correlation function (ACF) and the partial autocorrelation function (PACF) of the data are investigated with the goal of looking for the most appropriate time series model. 
In MTS, we applied the most commonly used method -- vector auto regression (VAR) on the datasets with and without the additional feature \textit{unemp\_rate} \cite{toda1994vector}, respectively.
The implementation of the UTS and MTS models are based on R and the Statsmodels library in Python 3, respectively.

\section{Results} \label{result}
\begin{figure}[htbp]
	\centering
	\includegraphics[width=3.3in]{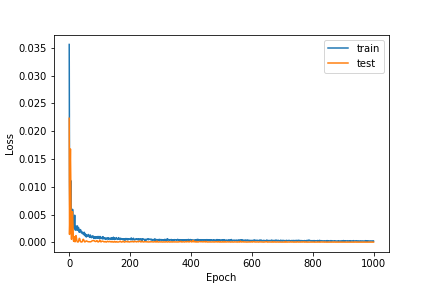}
	\caption{Loss on the Training and Test Sets for LSTM(1)}
	\label{lossLSTM1}
\end{figure}

\begin{figure}[htbp]
	\centering
	\includegraphics[width=3.3in]{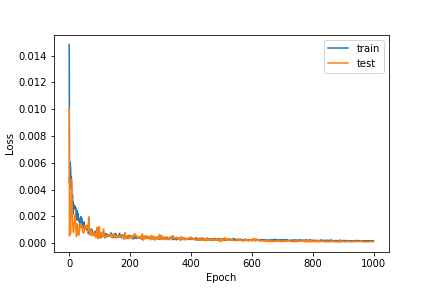}
	\caption{Loss on the Training and Test sets for LSTM(2)}
	\label{lossLSTM2}
\end{figure}

As discussed in Sections \ref{RNN} and \ref{TTS}, we first implement LSTM methodology and further compare its performance with traditional time series models. 
The critical step before the implementation of LSTM is the hyper-parameter tuning. 
By applying the trial and error approach via minimizing the loss on the test set, we finalized the values of the hyper-parameters as follows. 
The value of `number of nodes' is set to 70. 
`batch\_size' is set to 50 via trying different values ranging from 10 to 100 with a step of 10.
`number of epochs' is selected as 1000 to ensure the convergence of the algorithm.
Figures \ref{lossLSTM1} and \ref{lossLSTM2} show the changing of the loss value of training set and test set during each epoch for LSTM(1) and LSTM(2), respectively. 
LSTM(2) shows a smaller loss than LSTM(1) during the initial training stage but finally, the training process converges on both models.

On the other hand, the critical step before the implementation of traditional time series models is to ensure the stationary of the data.
The dataset is taken the first difference by making it stationary.
From the ACF and PACF plot, we see that autocorrelation decaying towards zero while PACF plot cuts off quickly towards zero. 
Therefore, for UTS, we only keep autoregressive component and have fitted traditional autoregressive models with order $p$ (i.e., AR($p$)) while the value of $p$ ranges from 1 to 3 in our study. 
The optimal value of $p$ in the AR model is identified as the one that generates the lowest Bayesian information criterion (BIC) value \cite{mousavi2008epileptic}.
Results show that AR(2) produces the lowest BIC values among the three AR models we compared (including AR(1), AR(2), and AR(3)). 
For MTS, it is interesting to find that VAR models perform much worse than AR(2) with respect to BIC, no matter whether the macroeconomic feature \textit{unemp\_rate} is used or not. 
Therefore, AR(2) is selected as the appropriate traditional time series model based on the P2P data in this study.

\begin{figure*}[htbp]
	\centering
	\includegraphics[width=4.7in]{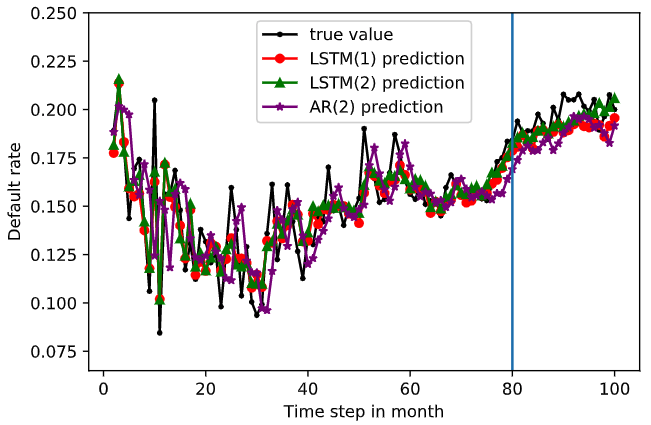}
	\caption{Predicted Trend of the Default Rate along with the True Trend from LSTM(1), LSTM(2), and AR(2)}
	\label{modelcompare}
\end{figure*}

Figure \ref{modelcompare} shows the predicted trend of \textit{default\_rate} from October 2007 to January 2016 along with the true trend using LSTM(1), LSTM(2), and AR(2) respectively.
The trend on the left of the vertical line is generated using the training data (i.e., data from October 2007 to May 2014) while the trend on the right is based on the test set (i.e., data from June 2014 to January 2016). 
We see that the predicted trend generated by LSTM(1) and LSTM(2) is very similar. 
Moreover, both LSTM models can capture the default trend very well.
However, there is an obvious delay in AR(2) in the ability of detecting the change. 
We further compare the RMSE values of the three models and the result is shown in Table \ref{RMSEcompare}.
AR(2) gives a much higher RMSE value on the testing set than that from either of the two LSTM models. 
Therefore, we conclude that LSTM shows its robustness in modeling the default rate of P2P market, no matter whether the macroeconomic feature \textit{unemp\_rate} is used or not. 
Furthermore, since LSTM(2) gives lower RMSE values on both training and testing set, indicating that incorporating the macroeconomic feature \textit{unemp\_rate} could further improve the model performance. 
All the above findings could further confirm the robustness of the LSTM method in modeling the sequential P2P data.


\begin{table}[htbp]
\renewcommand{\arraystretch}{1.3}
\caption{RMSE Comparison of the Three Models}
\label{RMSEcompare}
\centering
\begin{tabular}{|p{25mm}|p{25mm}|p{25mm}|}

			\hline
			Model & Training Set & Testing Set\\
			\hline
			LSTM(1)  & 0.013 & 0.010 \\
		    LSTM(2)  & 0.011 & 0.007 \\
		    AR(2)  & 0.019 & 0.021 \\
				\hline
		\end{tabular}
\end{table}




\section{Conclusion} \label{conclusion}
In this study, we aim to explore the monthly trend of the default rate on the aggregative level in the P2P lending market in the US. 
LSTM algorithm is first employed as a technique to model the sequential P2P transaction data. 
Considering the effect of the macroeconomic factor on the P2P market, we incorporate the unemployment rate (i.e. \textit{unemp\_rate}) as an additional predictor.
The result shows that although seldomly used in the P2P market, LSTM is a good alternative and even a more powerful tool to model the P2P transaction data compared to traditional time series models.
It is also demonstrated that adding \textit{unemp\_rate} could improve the LSTM performance by decreasing RMSE on both the training and the testing datasets. 
Different from previous studies that focus on modeling default risk at the individual level, our study provides a more comprehensive analysis of the P2P market by sequentially modeling the risk at the aggregative level.
Therefore, our study successfully broadens the application of the LSTM algorithm in the P2P market. 
Furthermore, our findings provide a good reference for investors to understand the entire status of the P2P market, especially the monthly trend of the default rate on an aggregative level. 
This is very critical in making their future investment strategies.

\bibliographystyle{ACM-Reference-Format}
\bibliography{sample-base}


\begin{thebibliography}{36}


\ifx \showCODEN    \undefined \def \showCODEN     #1{\unskip}     \fi
\ifx \showDOI      \undefined \def \showDOI       #1{#1}\fi
\ifx \showISBNx    \undefined \def \showISBNx     #1{\unskip}     \fi
\ifx \showISBNxiii \undefined \def \showISBNxiii  #1{\unskip}     \fi
\ifx \showISSN     \undefined \def \showISSN      #1{\unskip}     \fi
\ifx \showLCCN     \undefined \def \showLCCN      #1{\unskip}     \fi
\ifx \shownote     \undefined \def \shownote      #1{#1}          \fi
\ifx \showarticletitle \undefined \def \showarticletitle #1{#1}   \fi
\ifx \showURL      \undefined \def \showURL       {\relax}        \fi
\providecommand\bibfield[2]{#2}
\providecommand\bibinfo[2]{#2}
\providecommand\natexlab[1]{#1}
\providecommand\showeprint[2][]{arXiv:#2}

\bibitem[\protect\citeauthoryear{Bachmann, Becker, Buerckner, Hilker, Kock,
  Lehmann, Tiburtius, and Funk}{Bachmann et~al\mbox{.}}{2011}]%
        {bachmann2011online}
\bibfield{author}{\bibinfo{person}{A. Bachmann}, \bibinfo{person}{A. Becker},
  \bibinfo{person}{D. Buerckner}, \bibinfo{person}{M. Hilker},
  \bibinfo{person}{F. Kock}, \bibinfo{person}{M. Lehmann}, \bibinfo{person}{P.
  Tiburtius}, {and} \bibinfo{person}{B. Funk}.}
  \bibinfo{year}{2011}\natexlab{}.
\newblock \showarticletitle{Online Peer-to-peer Lending-a Literature Review}.
\newblock \bibinfo{journal}{\emph{Journal of Internet Banking and Commerce}}
  \bibinfo{volume}{16}, \bibinfo{number}{2} (\bibinfo{year}{2011}),
  \bibinfo{pages}{1}.
\newblock


\bibitem[\protect\citeauthoryear{Chen and Han}{Chen and Han}{1970}]%
        {chen1970comparative}
\bibfield{author}{\bibinfo{person}{D. Chen} {and} \bibinfo{person}{C. Han}.}
  \bibinfo{year}{1970}\natexlab{}.
\newblock \showarticletitle{A Comparative Study of Online P2P Lending in the
  USA and China}.
\newblock \bibinfo{journal}{\emph{The Journal of Internet Banking and
  Commerce}} \bibinfo{volume}{17}, \bibinfo{number}{2} (\bibinfo{year}{1970}),
  \bibinfo{pages}{12--15}.
\newblock


\bibitem[\protect\citeauthoryear{Emekter, Tu, Jirasakuldech, and Lu}{Emekter
  et~al\mbox{.}}{2015}]%
        {emekter2015evaluating}
\bibfield{author}{\bibinfo{person}{R. Emekter}, \bibinfo{person}{Y. Tu},
  \bibinfo{person}{B. Jirasakuldech}, {and} \bibinfo{person}{M. Lu}.}
  \bibinfo{year}{2015}\natexlab{}.
\newblock \showarticletitle{Evaluating Credit Risk and Loan Performance in
  Online Peer-to-Peer (P2P) Lending}.
\newblock \bibinfo{journal}{\emph{Applied Economics}} \bibinfo{volume}{47},
  \bibinfo{number}{1} (\bibinfo{year}{2015}), \bibinfo{pages}{54--70}.
\newblock


\bibitem[\protect\citeauthoryear{Foo, Lim, and Wong}{Foo et~al\mbox{.}}{2017}]%
        {foo2017macroeconomics}
\bibfield{author}{\bibinfo{person}{J. Foo}, \bibinfo{person}{L. Lim}, {and}
  \bibinfo{person}{K. Wong}.} \bibinfo{year}{2017}\natexlab{}.
\newblock \showarticletitle{Macroeconomics and Fintech: Uncovering Latent
  Macroeconomic Effects on Peer-to-peer Lending}.
\newblock \bibinfo{journal}{\emph{arXiv preprint arXiv:1710.11283}}
  (\bibinfo{year}{2017}).
\newblock


\bibitem[\protect\citeauthoryear{Graves, Jaitly, and Mohamed}{Graves
  et~al\mbox{.}}{2013a}]%
        {graves2013hybrid}
\bibfield{author}{\bibinfo{person}{A. Graves}, \bibinfo{person}{N. Jaitly},
  {and} \bibinfo{person}{A. Mohamed}.} \bibinfo{year}{2013}\natexlab{a}.
\newblock \showarticletitle{Hybrid Speech Recognition with Deep Bidirectional
  LSTM}. In \bibinfo{booktitle}{\emph{Automatic Speech Recognition and
  Understanding (ASRU), 2013 IEEE Workshop on}}. IEEE,
  \bibinfo{address}{Olomouc, Czech Republic}, \bibinfo{pages}{273--278}.
\newblock


\bibitem[\protect\citeauthoryear{Graves, Mohamed, and Hinton}{Graves
  et~al\mbox{.}}{2013b}]%
        {graves2013speech}
\bibfield{author}{\bibinfo{person}{A. Graves}, \bibinfo{person}{A. Mohamed},
  {and} \bibinfo{person}{G. Hinton}.} \bibinfo{year}{2013}\natexlab{b}.
\newblock \showarticletitle{Speech Recognition with Deep Recurrent Neural
  Networks}. In \bibinfo{booktitle}{\emph{2013 IEEE international conference on
  acoustics, speech and signal processing}}. IEEE, \bibinfo{address}{Vancouver,
  Canada}, \bibinfo{pages}{6645--6649}.
\newblock


\bibitem[\protect\citeauthoryear{Hammer and Steil}{Hammer and Steil}{2002}]%
        {hammer2002tutorial}
\bibfield{author}{\bibinfo{person}{B. Hammer} {and} \bibinfo{person}{J.
  Steil}.} \bibinfo{year}{2002}\natexlab{}.
\newblock \showarticletitle{Tutorial: Perspectives on Learning with RNNs}. In
  \bibinfo{booktitle}{\emph{Proc. ESANN}}. \bibinfo{address}{Bruges, Belgium},
  \bibinfo{pages}{357--368}.
\newblock


\bibitem[\protect\citeauthoryear{Hidasi, Karatzoglou, Baltrunas, and
  Tikk}{Hidasi et~al\mbox{.}}{2015}]%
        {hidasi2015session}
\bibfield{author}{\bibinfo{person}{B. Hidasi}, \bibinfo{person}{A.
  Karatzoglou}, \bibinfo{person}{L. Baltrunas}, {and} \bibinfo{person}{D.
  Tikk}.} \bibinfo{year}{2015}\natexlab{}.
\newblock \showarticletitle{Session-based Recommendations with Recurrent Neural
  Networks}.
\newblock \bibinfo{journal}{\emph{arXiv preprint arXiv:1511.06939}}
  (\bibinfo{year}{2015}).
\newblock


\bibitem[\protect\citeauthoryear{Ho, Xie, and Goh}{Ho et~al\mbox{.}}{2002}]%
        {ho2002comparative}
\bibfield{author}{\bibinfo{person}{S. Ho}, \bibinfo{person}{M. Xie}, {and}
  \bibinfo{person}{T. Goh}.} \bibinfo{year}{2002}\natexlab{}.
\newblock \showarticletitle{A Comparative Study of Neural Network and
  Box-Jenkins ARIMA Modeling in Time Series Prediction}.
\newblock \bibinfo{journal}{\emph{Computers \& Industrial Engineering}}
  \bibinfo{volume}{42}, \bibinfo{number}{2-4} (\bibinfo{year}{2002}),
  \bibinfo{pages}{371--375}.
\newblock


\bibitem[\protect\citeauthoryear{Hochreiter and Schmidhuber}{Hochreiter and
  Schmidhuber}{1997}]%
        {hochreiter1997lstm}
\bibfield{author}{\bibinfo{person}{S. Hochreiter} {and} \bibinfo{person}{J.
  Schmidhuber}.} \bibinfo{year}{1997}\natexlab{}.
\newblock \showarticletitle{LSTM Can Solve Hard Long Time Lag Problems}. In
  \bibinfo{booktitle}{\emph{Advances in neural information processing
  systems}}. \bibinfo{pages}{473--479}.
\newblock


\bibitem[\protect\citeauthoryear{Jagannatha and Yu}{Jagannatha and Yu}{2016}]%
        {jagannatha2016bidirectional}
\bibfield{author}{\bibinfo{person}{A. Jagannatha} {and} \bibinfo{person}{H.
  Yu}.} \bibinfo{year}{2016}\natexlab{}.
\newblock \showarticletitle{Bidirectional RNN for Medical Event Detection in
  Electronic Health Records}. In \bibinfo{booktitle}{\emph{Proceedings of the
  conference. Association for Computational Linguistics. North American
  Chapter. Meeting}}, Vol.~\bibinfo{volume}{2016}. NIH Public Access,
  \bibinfo{address}{San Diego, California}, \bibinfo{pages}{473}.
\newblock


\bibitem[\protect\citeauthoryear{Jain, Mao, and Mohiuddin}{Jain
  et~al\mbox{.}}{1996}]%
        {jain1996artificial}
\bibfield{author}{\bibinfo{person}{A. Jain}, \bibinfo{person}{J. Mao}, {and}
  \bibinfo{person}{K. Mohiuddin}.} \bibinfo{year}{1996}\natexlab{}.
\newblock \showarticletitle{Artificial Neural Networks: A tutorial}.
\newblock \bibinfo{journal}{\emph{Computer}} \bibinfo{number}{3}
  (\bibinfo{year}{1996}), \bibinfo{pages}{31--44}.
\newblock


\bibitem[\protect\citeauthoryear{Jakubik et~al\mbox{.}}{Jakubik
  et~al\mbox{.}}{2007}]%
        {jakubik2007macroeconomic}
\bibfield{author}{\bibinfo{person}{P. Jakubik} {et~al\mbox{.}}}
  \bibinfo{year}{2007}\natexlab{}.
\newblock \showarticletitle{Macroeconomic Environment and Credit Risk}.
\newblock \bibinfo{journal}{\emph{Czech Journal of Economics and Finance
  (Finance a uver)}} \bibinfo{volume}{57}, \bibinfo{number}{1-2}
  (\bibinfo{year}{2007}), \bibinfo{pages}{60--78}.
\newblock


\bibitem[\protect\citeauthoryear{Jurgovsky, Granitzer, Ziegler, Calabretto,
  Portier, He-Guelton, and Caelen}{Jurgovsky et~al\mbox{.}}{2018}]%
        {jurgovsky2018sequence}
\bibfield{author}{\bibinfo{person}{J. Jurgovsky}, \bibinfo{person}{M.
  Granitzer}, \bibinfo{person}{K. Ziegler}, \bibinfo{person}{S. Calabretto},
  \bibinfo{person}{P. Portier}, \bibinfo{person}{L. He-Guelton}, {and}
  \bibinfo{person}{O. Caelen}.} \bibinfo{year}{2018}\natexlab{}.
\newblock \showarticletitle{Sequence Classification for Credit-card Fraud
  Detection}.
\newblock \bibinfo{journal}{\emph{Expert Systems with Applications}}
  \bibinfo{volume}{100} (\bibinfo{year}{2018}), \bibinfo{pages}{234--245}.
\newblock


\bibitem[\protect\citeauthoryear{Kim and Cho}{Kim and Cho}{2019}]%
        {kim2019predicting}
\bibfield{author}{\bibinfo{person}{J. Kim} {and} \bibinfo{person}{S. Cho}.}
  \bibinfo{year}{2019}\natexlab{}.
\newblock \showarticletitle{Predicting Repayment of Borrows in Peer-to-peer
  Social Lending with Deep Dense Convolutional Network}.
\newblock \bibinfo{journal}{\emph{Expert Systems}} (\bibinfo{year}{2019}),
  \bibinfo{pages}{e12403}.
\newblock


\bibitem[\protect\citeauthoryear{Le et~al\mbox{.}}{Le et~al\mbox{.}}{2015}]%
        {le2015tutorial}
\bibfield{author}{\bibinfo{person}{Q. Le} {et~al\mbox{.}}}
  \bibinfo{year}{2015}\natexlab{}.
\newblock \showarticletitle{A Tutorial on Deep Learning Part 2: Autoencoders,
  Convolutional Neural Networks and Recurrent Neural Networks}.
\newblock \bibinfo{journal}{\emph{Google Brain}} (\bibinfo{year}{2015}),
  \bibinfo{pages}{1--20}.
\newblock


\bibitem[\protect\citeauthoryear{Li, Long, Sun, Yang, and Li}{Li
  et~al\mbox{.}}{2018}]%
        {li2018overdue}
\bibfield{author}{\bibinfo{person}{X. Li}, \bibinfo{person}{X. Long},
  \bibinfo{person}{G. Sun}, \bibinfo{person}{G. Yang}, {and}
  \bibinfo{person}{H. Li}.} \bibinfo{year}{2018}\natexlab{}.
\newblock \showarticletitle{Overdue Prediction of Bank Loans Based on
  LSTM-SVM}. In \bibinfo{booktitle}{\emph{2018 IEEE SmartWorld, Ubiquitous
  Intelligence \& Computing, Advanced \& Trusted Computing, Scalable Computing
  \& Communications, Cloud \& Big Data Computing, Internet of People and Smart
  City Innovation (SmartWorld/SCALCOM/UIC/ATC/CBDCom/IOP/SCI)}}. IEEE,
  \bibinfo{address}{Guandong, China}, \bibinfo{pages}{1859--1863}.
\newblock


\bibitem[\protect\citeauthoryear{Lipton, Berkowitz, and Elkan}{Lipton
  et~al\mbox{.}}{2015}]%
        {lipton2015critical}
\bibfield{author}{\bibinfo{person}{Z. Lipton}, \bibinfo{person}{J. Berkowitz},
  {and} \bibinfo{person}{C. Elkan}.} \bibinfo{year}{2015}\natexlab{}.
\newblock \showarticletitle{A Critical Review of Recurrent Neural Networks for
  Sequence Learning}.
\newblock \bibinfo{journal}{\emph{arXiv preprint arXiv:1506.00019}}
  (\bibinfo{year}{2015}).
\newblock


\bibitem[\protect\citeauthoryear{Liu, Han, Y.Zhou, and Wang}{Liu
  et~al\mbox{.}}{2018}]%
        {liu2018lstm}
\bibfield{author}{\bibinfo{person}{L. Liu}, \bibinfo{person}{M. Han},
  \bibinfo{person}{Y.Zhou}, {and} \bibinfo{person}{Y. Wang}.}
  \bibinfo{year}{2018}\natexlab{}.
\newblock \showarticletitle{LSTM Recurrent Neural Networks for Influenza Trends
  Prediction}. In \bibinfo{booktitle}{\emph{International Symposium on
  Bioinformatics Research and Applications}}. Springer,
  \bibinfo{address}{Beijing, China}, \bibinfo{pages}{259--264}.
\newblock


\bibitem[\protect\citeauthoryear{Malekipirbazari and Aksakalli}{Malekipirbazari
  and Aksakalli}{2015}]%
        {malekipirbazari2015risk}
\bibfield{author}{\bibinfo{person}{M. Malekipirbazari} {and}
  \bibinfo{person}{V. Aksakalli}.} \bibinfo{year}{2015}\natexlab{}.
\newblock \showarticletitle{Risk Assessment in Social Lending via Random
  Forests}.
\newblock \bibinfo{journal}{\emph{Expert Systems with Applications}}
  \bibinfo{volume}{42}, \bibinfo{number}{10} (\bibinfo{year}{2015}),
  \bibinfo{pages}{4621--4631}.
\newblock


\bibitem[\protect\citeauthoryear{Marino, Amarasinghe, and Manic}{Marino
  et~al\mbox{.}}{2016}]%
        {marino2016building}
\bibfield{author}{\bibinfo{person}{D. Marino}, \bibinfo{person}{K.
  Amarasinghe}, {and} \bibinfo{person}{M. Manic}.}
  \bibinfo{year}{2016}\natexlab{}.
\newblock \showarticletitle{Building Energy Load Forecasting using Deep Neural
  Networks}. In \bibinfo{booktitle}{\emph{IECON 2016-42nd Annual Conference of
  the IEEE Industrial Electronics Society}}. IEEE, \bibinfo{address}{Florence,
  Italy}, \bibinfo{pages}{7046--7051}.
\newblock


\bibitem[\protect\citeauthoryear{Mariotto}{Mariotto}{2016}]%
        {mariotto2016competition}
\bibfield{author}{\bibinfo{person}{C. Mariotto}.}
  \bibinfo{year}{2016}\natexlab{}.
\newblock \showarticletitle{Competition for Lending in the Internet Era: The
  Case of Peer-to-Peer Lending Marketplaces in the USA}.
\newblock \bibinfo{journal}{\emph{Available at SSRN 2800998}}
  (\bibinfo{year}{2016}).
\newblock


\bibitem[\protect\citeauthoryear{Molina, Bhulai, Reader, Jeu, and
  Tjepkema}{Molina et~al\mbox{.}}{2018}]%
        {molina2018understanding}
\bibfield{author}{\bibinfo{person}{L. Molina}, \bibinfo{person}{S. Bhulai},
  \bibinfo{person}{V. Reader}, \bibinfo{person}{R. Jeu}, {and}
  \bibinfo{person}{J. Tjepkema}.} \bibinfo{year}{2018}\natexlab{}.
\newblock \showarticletitle{Understanding User Behavior in E-commerce with Long
  Short-term Memory (LSTM) and Autoencoders}.
\newblock  (\bibinfo{year}{2018}).
\newblock


\bibitem[\protect\citeauthoryear{Mousavi, Niknazar, and Vahdat}{Mousavi
  et~al\mbox{.}}{2008}]%
        {mousavi2008epileptic}
\bibfield{author}{\bibinfo{person}{S. Mousavi}, \bibinfo{person}{M. Niknazar},
  {and} \bibinfo{person}{B. Vahdat}.} \bibinfo{year}{2008}\natexlab{}.
\newblock \showarticletitle{Epileptic Seizure Detection using AR Model on EEG
  Signals}. In \bibinfo{booktitle}{\emph{2008 Cairo International Biomedical
  Engineering Conference}}. IEEE, \bibinfo{address}{Cairo, Egypt},
  \bibinfo{pages}{1--4}.
\newblock


\bibitem[\protect\citeauthoryear{Namvar, Siami, Rabhi, and Naderpour}{Namvar
  et~al\mbox{.}}{2018}]%
        {namvar2018credit}
\bibfield{author}{\bibinfo{person}{A. Namvar}, \bibinfo{person}{M. Siami},
  \bibinfo{person}{F. Rabhi}, {and} \bibinfo{person}{M. Naderpour}.}
  \bibinfo{year}{2018}\natexlab{}.
\newblock \showarticletitle{Credit Risk Prediction in an Imbalanced Social
  Lending Environment}.
\newblock \bibinfo{journal}{\emph{arXiv preprint arXiv:1805.00801}}
  (\bibinfo{year}{2018}).
\newblock


\bibitem[\protect\citeauthoryear{Potdar, Pardawala, and Pai}{Potdar
  et~al\mbox{.}}{2017}]%
        {potdar2017comparative}
\bibfield{author}{\bibinfo{person}{K. Potdar}, \bibinfo{person}{T. Pardawala},
  {and} \bibinfo{person}{C. Pai}.} \bibinfo{year}{2017}\natexlab{}.
\newblock \showarticletitle{A Comparative Study of Categorical Variable
  Encoding Techniques for Neural Network Classifiers}.
\newblock \bibinfo{journal}{\emph{International Journal of Computer
  Applications}} \bibinfo{volume}{175}, \bibinfo{number}{4}
  (\bibinfo{year}{2017}), \bibinfo{pages}{7--9}.
\newblock


\bibitem[\protect\citeauthoryear{Ren, Hu, Tai, Wang, Xu, Sun, and Yan}{Ren
  et~al\mbox{.}}{2016}]%
        {ren2016look}
\bibfield{author}{\bibinfo{person}{J. Ren}, \bibinfo{person}{Y. Hu},
  \bibinfo{person}{Y. Tai}, \bibinfo{person}{C. Wang}, \bibinfo{person}{L. Xu},
  \bibinfo{person}{W. Sun}, {and} \bibinfo{person}{Q. Yan}.}
  \bibinfo{year}{2016}\natexlab{}.
\newblock \showarticletitle{Look, Listen and Learn-A Multimodal LSTM for
  Speaker Identification.}. In \bibinfo{booktitle}{\emph{AAAI}}.
  \bibinfo{address}{Phoenix, Arizona USA}, \bibinfo{pages}{3581--3587}.
\newblock


\bibitem[\protect\citeauthoryear{S{\"o}gner}{S{\"o}gner}{2001}]%
        {sogner2001okun}
\bibfield{author}{\bibinfo{person}{L. S{\"o}gner}.}
  \bibinfo{year}{2001}\natexlab{}.
\newblock \showarticletitle{Okun's Law Does the Austrian Unemployment--GDP
  Relationship Exhibit Structural Breaks?}
\newblock \bibinfo{journal}{\emph{Empirical Economics}} \bibinfo{volume}{26},
  \bibinfo{number}{3} (\bibinfo{year}{2001}), \bibinfo{pages}{553--564}.
\newblock


\bibitem[\protect\citeauthoryear{Sundermeyer, Schl{\"u}ter, and
  Ney}{Sundermeyer et~al\mbox{.}}{2012}]%
        {sundermeyer2012lstm}
\bibfield{author}{\bibinfo{person}{M. Sundermeyer}, \bibinfo{person}{R.
  Schl{\"u}ter}, {and} \bibinfo{person}{H. Ney}.}
  \bibinfo{year}{2012}\natexlab{}.
\newblock \showarticletitle{LSTM Neural Networks for Language Modeling}. In
  \bibinfo{booktitle}{\emph{Thirteenth annual conference of the international
  speech communication association}}. \bibinfo{address}{Portland, Oregon}.
\newblock


\bibitem[\protect\citeauthoryear{Toda and Phillips}{Toda and Phillips}{1994}]%
        {toda1994vector}
\bibfield{author}{\bibinfo{person}{H. Toda} {and} \bibinfo{person}{P.
  Phillips}.} \bibinfo{year}{1994}\natexlab{}.
\newblock \showarticletitle{Vector Autoregression and Causality: a Theoretical
  overview and Simulation Study}.
\newblock \bibinfo{journal}{\emph{Econometric reviews}} \bibinfo{volume}{13},
  \bibinfo{number}{2} (\bibinfo{year}{1994}), \bibinfo{pages}{259--285}.
\newblock


\bibitem[\protect\citeauthoryear{Wang, Han, Liu, and Luo}{Wang
  et~al\mbox{.}}{2018}]%
        {wang2018deep}
\bibfield{author}{\bibinfo{person}{C. Wang}, \bibinfo{person}{D. Han},
  \bibinfo{person}{Q. Liu}, {and} \bibinfo{person}{S. Luo}.}
  \bibinfo{year}{2018}\natexlab{}.
\newblock \showarticletitle{A Deep Learning Approach for Credit Scoring of
  Peer-to-Peer Lending using Attention Mechanism LSTM}.
\newblock \bibinfo{journal}{\emph{IEEE Access}}  \bibinfo{volume}{7}
  (\bibinfo{year}{2018}), \bibinfo{pages}{2161--2168}.
\newblock


\bibitem[\protect\citeauthoryear{Wiese and Omlin}{Wiese and Omlin}{2009}]%
        {wiese2009credit}
\bibfield{author}{\bibinfo{person}{B. Wiese} {and} \bibinfo{person}{C. Omlin}.}
  \bibinfo{year}{2009}\natexlab{}.
\newblock \showarticletitle{Credit Card Transactions, Fraud Detection, and
  Machine Learning: Modelling time with LSTM Recurrent Neural Networks}.
\newblock In \bibinfo{booktitle}{\emph{Innovations in neural information
  paradigms and applications}}. \bibinfo{publisher}{Springer},
  \bibinfo{pages}{231--268}.
\newblock


\bibitem[\protect\citeauthoryear{Xiong and Shang}{Xiong and Shang}{2017}]%
        {xiong2017detrended}
\bibfield{author}{\bibinfo{person}{H. Xiong} {and} \bibinfo{person}{P. Shang}.}
  \bibinfo{year}{2017}\natexlab{}.
\newblock \showarticletitle{Detrended Fluctuation Analysis of Multivariate Time
  Series}.
\newblock \bibinfo{journal}{\emph{Communications in Nonlinear Science and
  Numerical Simulation}}  \bibinfo{volume}{42} (\bibinfo{year}{2017}),
  \bibinfo{pages}{12--21}.
\newblock


\bibitem[\protect\citeauthoryear{Ye, Dong, and Ma}{Ye et~al\mbox{.}}{2018}]%
        {ye2018loan}
\bibfield{author}{\bibinfo{person}{X. Ye}, \bibinfo{person}{L. Dong}, {and}
  \bibinfo{person}{D. Ma}.} \bibinfo{year}{2018}\natexlab{}.
\newblock \showarticletitle{Loan Evaluation in P2P Lending based on Random
  Forest Optimized by Genetic Algorithm with Profit Score}.
\newblock \bibinfo{journal}{\emph{Electronic Commerce Research and
  Applications}}  \bibinfo{volume}{32} (\bibinfo{year}{2018}),
  \bibinfo{pages}{23--36}.
\newblock


\bibitem[\protect\citeauthoryear{Zhang, Zhao, Xue, Chen, Ma, and Zhou}{Zhang
  et~al\mbox{.}}{2017}]%
        {zhang2017time}
\bibfield{author}{\bibinfo{person}{J. Zhang}, \bibinfo{person}{Z. Zhao},
  \bibinfo{person}{Y. Xue}, \bibinfo{person}{Z. Chen}, \bibinfo{person}{X. Ma},
  {and} \bibinfo{person}{Q. Zhou}.} \bibinfo{year}{2017}\natexlab{}.
\newblock \showarticletitle{Time Series Analysis}.
\newblock \bibinfo{journal}{\emph{Handbook of Medical Statistics}}
  \bibinfo{volume}{269} (\bibinfo{year}{2017}).
\newblock


\bibitem[\protect\citeauthoryear{Zohrevand, Gl{\"a}sser, Tayebi, Shahir,
  Shirmaleki, and Shahir}{Zohrevand et~al\mbox{.}}{2017}]%
        {zohrevand2017deep}
\bibfield{author}{\bibinfo{person}{Z. Zohrevand}, \bibinfo{person}{U.
  Gl{\"a}sser}, \bibinfo{person}{M. Tayebi}, \bibinfo{person}{H. Shahir},
  \bibinfo{person}{M. Shirmaleki}, {and} \bibinfo{person}{A. Shahir}.}
  \bibinfo{year}{2017}\natexlab{}.
\newblock \showarticletitle{Deep Learning based Forecasting of Critical
  Infrastructure Data}. In \bibinfo{booktitle}{\emph{Proceedings of the 2017
  ACM on Conference on Information and Knowledge Management}}. ACM,
  \bibinfo{address}{Singapore Singapore}, \bibinfo{pages}{1129--1138}.
\newblock


\end{thebibliography}

%

\end{document}